\documentclass[journal,transmag]{IEEEtran}

\usepackage[utf8]{inputenc}
\usepackage{todonotes}
\usepackage{amsfonts}
\usepackage{tabularx}
\usepackage{array, booktabs, makecell}
\usepackage{siunitx}
\usepackage{xspace}
\usepackage{tablefootnote}
\usepackage{lipsum}
\usepackage{amsmath,amssymb}
\usepackage{hyperref}
\usepackage[font=small,labelfont=bf,tableposition=top]{caption}
\usepackage{multirow}

\newcommand{\simname}{SUMMIT\xspace}
\newcommand{\algname}{Context-POMDP\xspace}
\newcommand{\modelname}{GAMMA\xspace}

\newcommand{\secref}[1]{Section~\ref{#1}}
\renewcommand{\eqref}[1]{Eqn.~(\ref{#1})}
\newcommand{\figref}[1]{Fig.~\ref{#1}}

\newcommand{\tabref}[1]{Table~\ref{#1}}
\newcommand{\subfig}[1]{\textit{#1}}

\newcommand{\ie}{\textrm{i.e.}}
\newcommand{\eg}{\textrm{e.g.}}
\newcommand{\etc}{\textrm{etc.}}

\newcommand{\Fbox}[1]{\setlength{\fboxrule}{0pt}\setlength{\fboxsep}{0pt}\fbox{#1}}

\newcommand{\internalstates}{human inner states\xspace}

\newcommand{\vo}{\ensuremath{\mathrm{VO}\xspace}} 
\newcommand{\vel}{\ensuremath{v\xspace}} 
\newcommand{\velu}{\ensuremath{u\xspace}} 
\newcommand{\veln}{\ensuremath{n\xspace}} 
\newcommand{\pose}{\ensuremath{\mathrm{\mathbf{p\xspace}}}} 
\newcommand{\timewin}{\ensuremath{\tau\xspace}} 
\newcommand{\responsibility}{\ensuremath{\gamma\xspace}} 

\newcommand{\attfun}{\ensuremath{\mathrm{Att}\xspace}}
\newcommand{\kinfun}{\ensuremath{f\xspace}}
\newcommand{\kinset}{\ensuremath{\mathrm{K}\xspace}}
\newcommand{\approxkinset}{\ensuremath{\hat{\mathrm{K}}\xspace}}
\newcommand{\geoset}{\ensuremath{\mathrm{G}\xspace}}
\newcommand{\conset}{\ensuremath{\mathrm{C}\xspace}}
\newcommand{\maxbounderror}{\ensuremath{\mathrm{\varepsilon_{\mathrm{max}}}\xspace}}

\newcommand{\norm}[1]{\left\lVert#1\right\rVert}
\DeclareMathOperator*{\argmin}{arg\,min} 
\DeclareSIUnit\Ms{m/s}

\begin{document}

\title{\simname: Simulating Autonomous Driving in \\Massive Mixed Urban Traffic}


\author{\IEEEauthorblockN{Yuanfu Luo\IEEEauthorrefmark{*,1},
Panpan Cai\IEEEauthorrefmark{*,1,\dag},
Yiyuan Lee\IEEEauthorrefmark{1}, and 
David Hsu\IEEEauthorrefmark{1},~\IEEEmembership{Fellow,~IEEE}}
\IEEEauthorblockA{\IEEEauthorrefmark{1}School of Computing, National University of Singapore, 117417 Singapore}
\IEEEauthorblockA{\IEEEauthorrefmark{*}The authors have contributed equally.}
\IEEEauthorblockA{\IEEEauthorrefmark{\dag}Corresponding author: P. Cai (email: dcscaip@nus.edu.sg).}
}

\IEEEtitleabstractindextext{%
\begin{abstract}
Autonomous driving in an unregulated urban crowd is an outstanding challenge,
especially, in the presence of many aggressive, high-speed traffic
participants.  This paper presents \simname, a high-fidelity simulator that
facilitates the development and testing of crowd-driving algorithms. \simname
simulates dense, unregulated urban traffic at any worldwide locations as supported by the OpenStreetMap. The core of \simname is a multi-agent motion model, \modelname, that models the behaviours of heterogeneous traffic agents, and a real-time POMDP planner, \algname, that serves as a driving expert. \simname is built as an
extension of CARLA and inherits from it the physical and visual realism for
autonomous driving simulation.  \simname supports a wide range of
applications, including perception, vehicle control or planning, and end-to-end
learning. We validate the realism of our motion model using its traffic motion prediction accuracy on various real-world data sets. We also provide several real-world benchmark scenarios to show that \simname simulates complex, realistic traffic behaviors, and \algname drives safely and efficiently in challenging crowd-driving settings.

\end{abstract}

}

\twocolumn[{%
\renewcommand\twocolumn[1][]{#1}%

\maketitle
\IEEEdisplaynontitleabstractindextext
\IEEEpeerreviewmaketitle

\begin{center}
\centering                                                              
\begin{tabular}{ccc}
\Fbox{\includegraphics[width=0.3\textwidth]{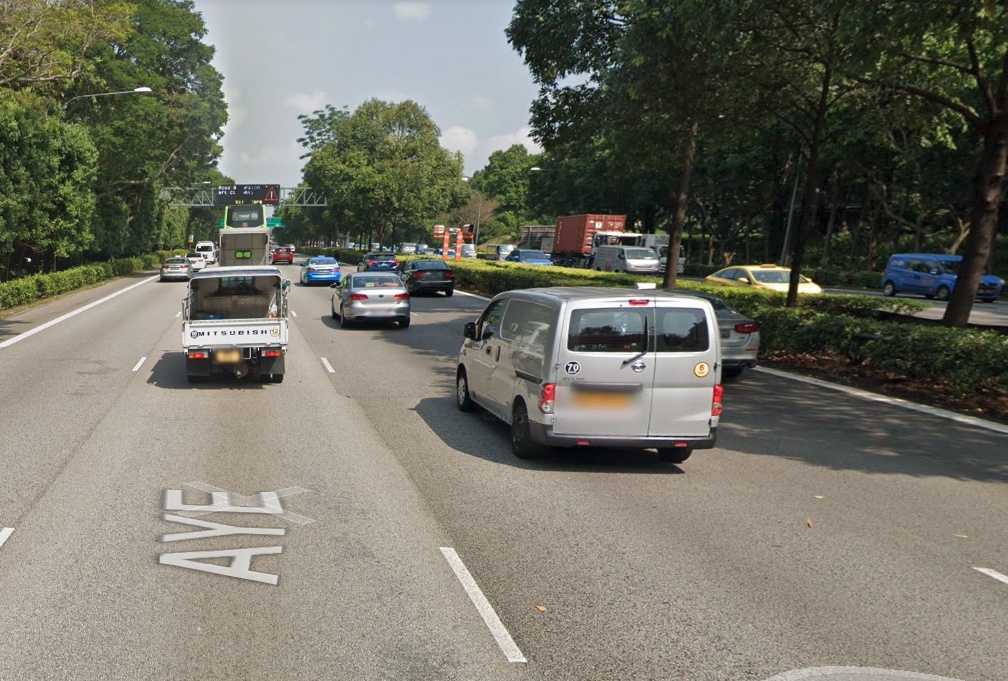}} & \Fbox{\includegraphics[width=0.3\textwidth]{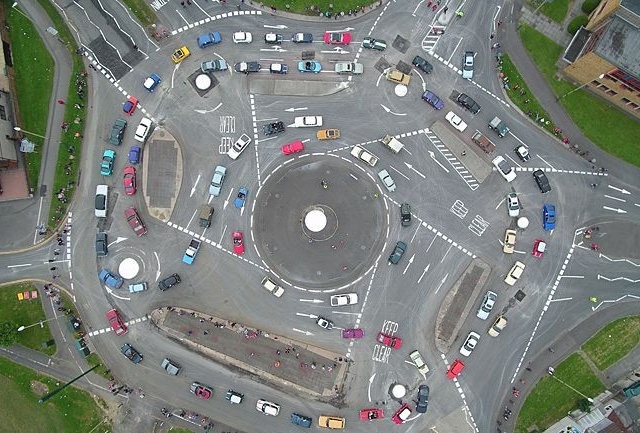}} & 
\Fbox{\includegraphics[width=0.3\textwidth]{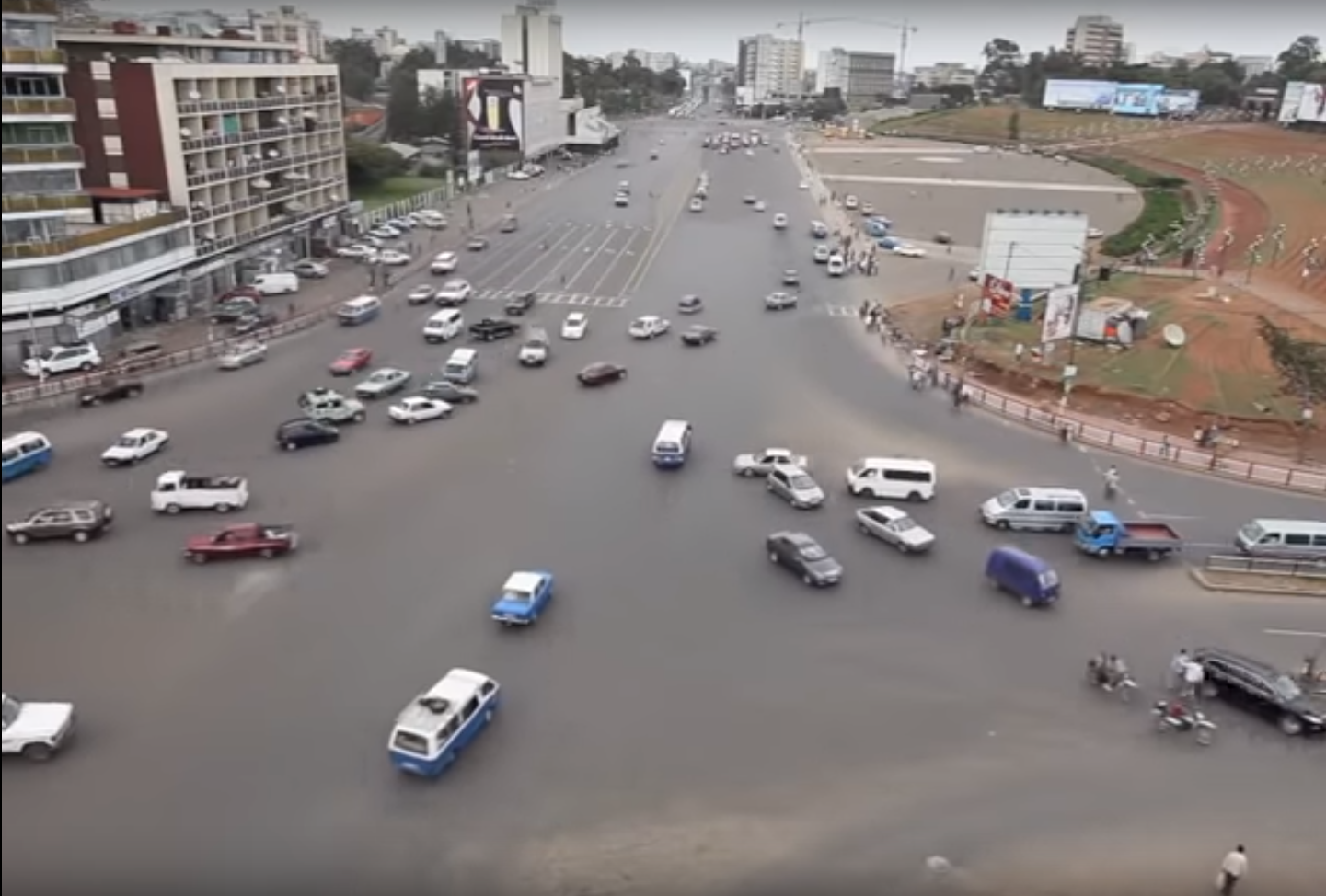}} \\
(\subfig{a}) Singapore-Highway & (\subfig{b}) Magic-Roundabout & (\subfig{c}) Meskel-Intersection\\    
\Fbox{\includegraphics[width=0.3\textwidth]{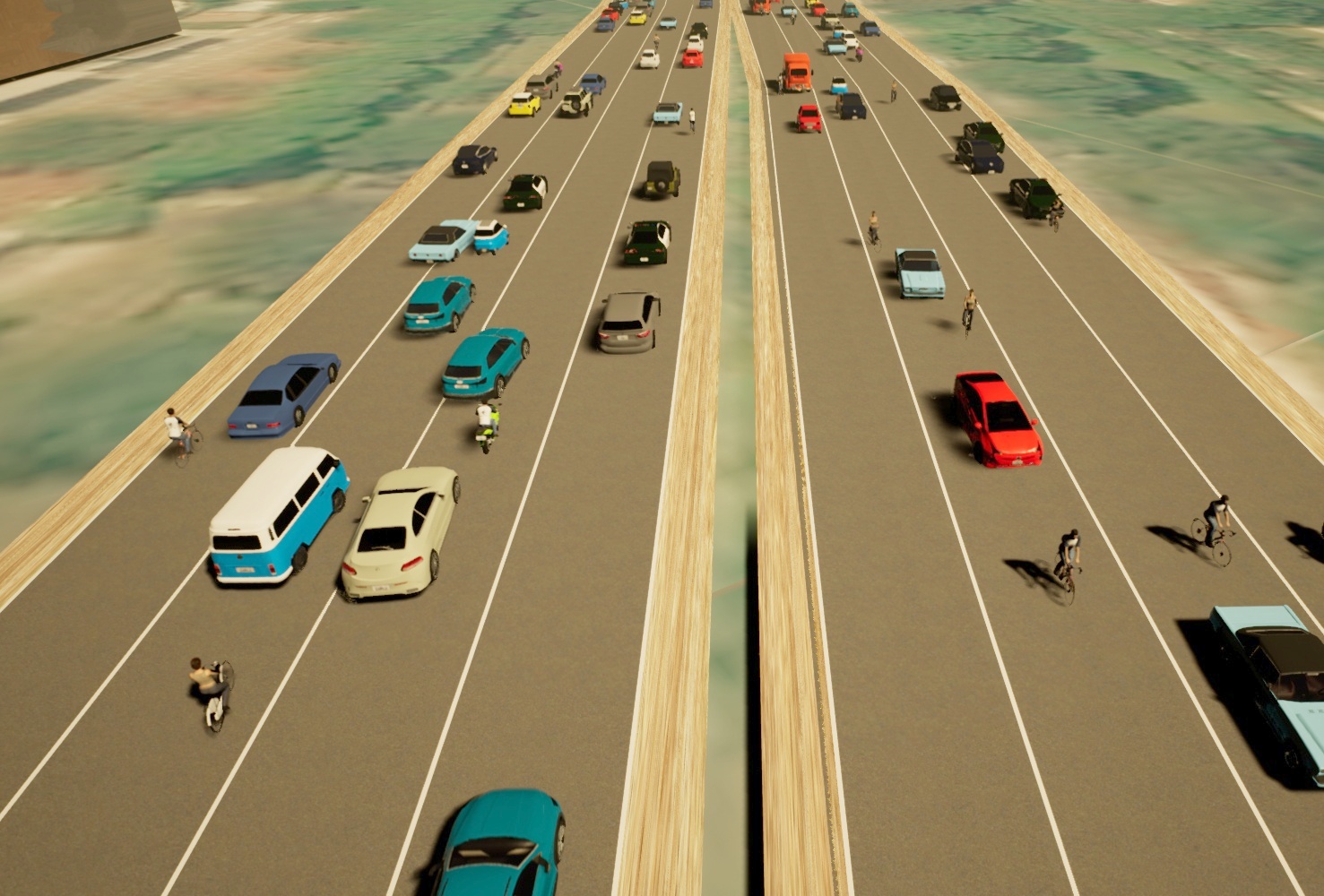}} &  \Fbox{\includegraphics[width=0.3\textwidth]{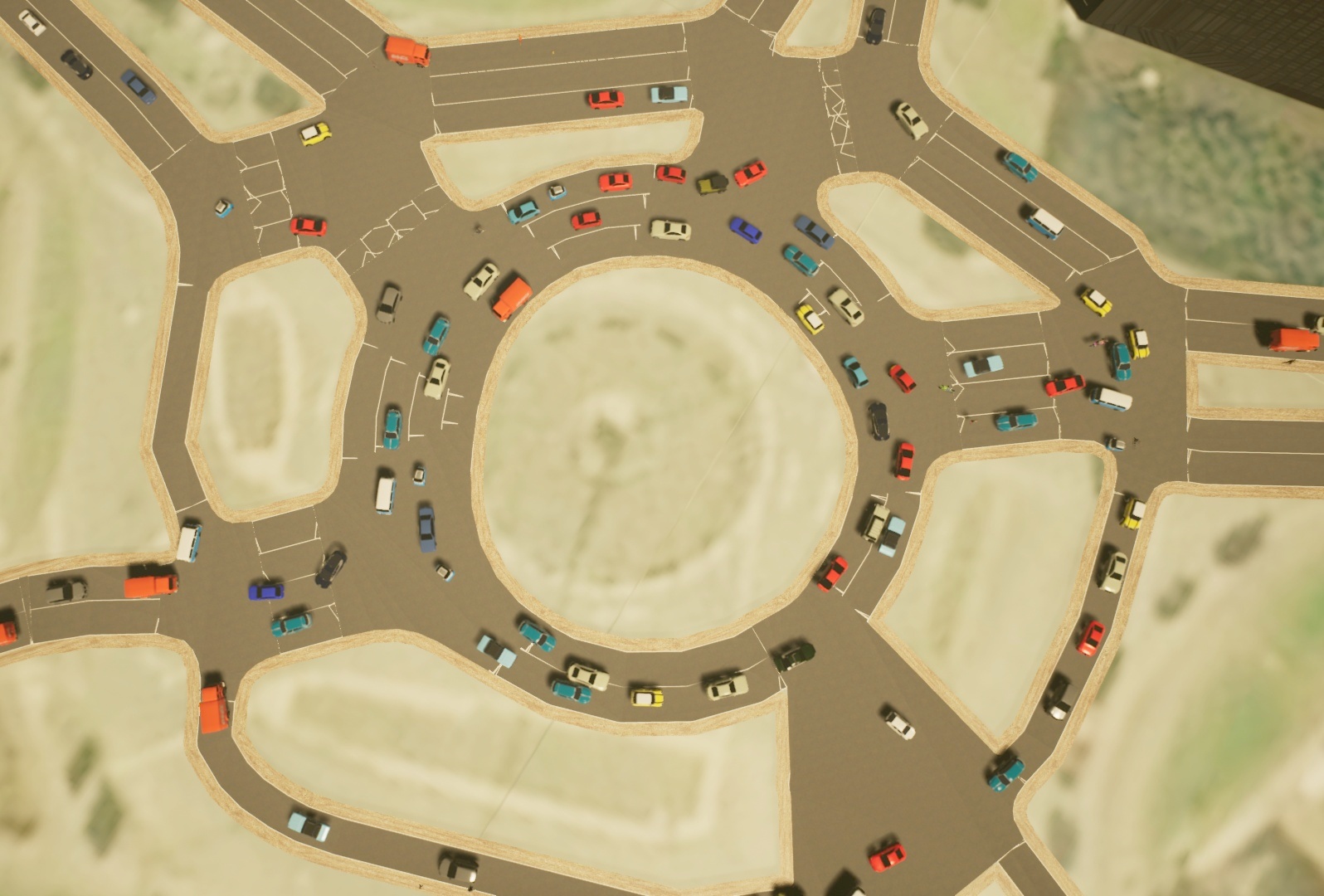}} &  \Fbox{\includegraphics[width=0.3\textwidth]{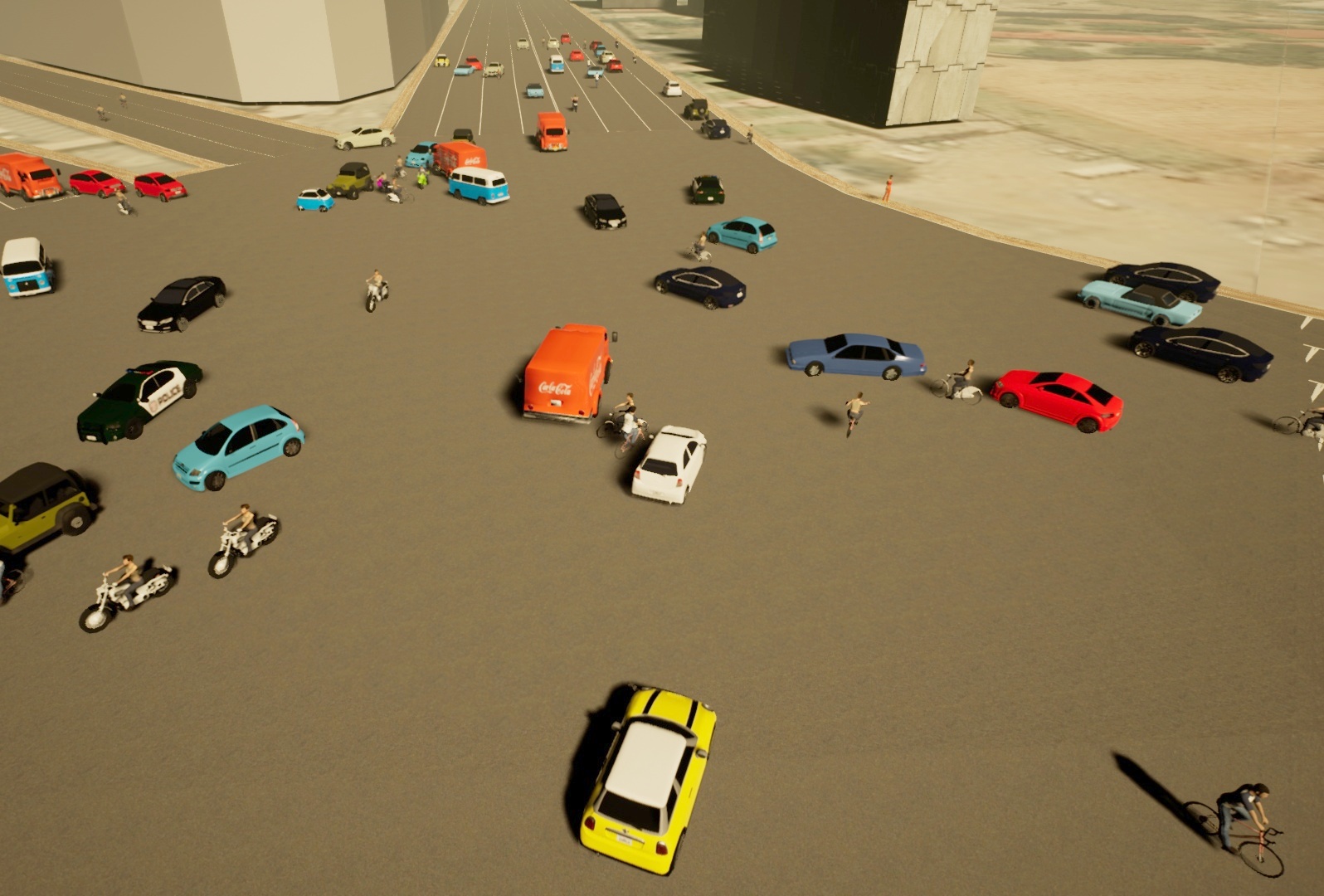}}  \\
(\subfig{d}) Singapore-Highway in simulation & (\subfig{e}) Magic-Roundabout in simulation & (\subfig{f}) Meskel-Intersection in simulation\\    
\end{tabular}  
\captionof{figure}{ Real-world driving scenes and corresponding scenes in simulation.
}        
\label{fig:benchmarks}     
\end{center} 
}]


\section{Introduction}
\IEEEPARstart{T}{he} vision of using autonomous driving to improve the safety and convenience of our daily life is coming closer. However, driving through a crowded urban traffic, especially in uncontrolled roads or unsignalised intersections (\figref{fig:benchmarks}), remains an open problem. 
Technical challenges for \textit{crowd-driving} come from both the complexity of crowd behaviors and map environments. 
Urban traffic can be extremely chaotic due to the \textit{unregulated} behaviors - drivers disregard traffic rules and drive aggressively, and intensive \textit{interactions} - collision avoidance, competition, and negotiation. Traffic agents are also \emph{heterogeneous}, \ie, having different geometry, kinematics, and dynamics. Human participants also have different \textit{inner states} or behavioral modes - intending at different routes, being attentive or distracted, responsible or careless, \etc. 
The complexity is further added-up by versatile urban roads: multi-lane roads, intersections, roundabouts - they generate very different crowd behaviors. 

 
High-quality data for tackling crowd-driving are, however, difficult and expensive to acquire, due to the cost of devices, regulations and safety constraints. Although there are publicly available data sets like  KITTI \cite{KITTI}, BDD100K \cite{BDD100K}, Oxford RobotCar \cite{Oxford}, \etc, providing real-world driving data with rich sensor inputs, these data are not \emph{interactive}, \ie, one cannot model the reactions of exo-agents to the robot's decisions. However, interactive data are extremely important, \eg, for planners and RL algorithms to reliably estimate their task performance. 
Driving simulators offer the capability of generating imaginary interactive driving data. However, existing driving simulators have not captured the full complexity of unregulated urban crowds and real-world maps. They are thus insufficient for testing or training robust crowd-driving algorithms. We aim to fill this gap.

We develop a new simulator, \simname, that generates high-fidelity, interactive data for unregulated, dense urban traffic on complex real-world maps. 
\simname fetches real-world maps from online sources to provide a virtually unlimited source of complex environments. Given arbitrary locations, \simname automatically generates crowds of heterogeneous traffic agents and simulates their local interactions using a realistic motion model. Global traffic conditions are constructed using road contexts to guide the crowd behaviours topologically and geometrically. We implemented \simname based on CARLA \cite{CARLA} to leverage the high-fidelity physics, rendering, ans sensors. Through a python-based API, \simname reveals rich sensor data, semantic information, and road contexts to external algorithms, enabling the application in a wide range of fields such as perception, vehicle control and planning, end-to-end learning, etc. 

The core of \simname is our general agent motion prediction model for autonomous driving (\modelname).
\modelname encodes four key factors that govern traffic agents' motions: \emph{kinematics}, \emph{geometry} (collision avoidance), \emph{intention} (destination), and \emph{road context}. We formulate traffic motion modelling as constrained optimization in the velocity space of agents. Constraints of the problem encode kinematic, collision avoidance, and road structures; Objective of the problem is to conduct the intended motion efficiently, \eg, driving along an intended lane at a desired speed. 


\modelname significantly extends existing Reciprocal Velocity Obstacle (RVO) frameworks \cite{van2011reciprocal} in the following aspects. It uses an efficient numerical way to enforce kinematic constraints, thus avoid detailed modeling of kinematics. Second, we support polygons-based representation, which offers tighter fit to traffic agents than typical disk representations, especially for long or narrow vehicles like trucks and bicycles. Lastly, \modelname explicitly models \internalstates{}, making the model highly variable. By applying Bayesian inference on uncertain \internalstates{} and conditioning \modelname on the posterior distribution, the model can generate very accurate predictions as validated on real-world datasets and thus empower online planning under uncertainty.


We further provide a new crowd-driving algorithm, \algname, to serve as a driving expert in \simname. We formulate crowd-driving as a partially observable Markov decision process (POMDP) which treats the \internalstates{} as hidden variables and capture the uncertainty using beliefs. Near-optimal planning is achieved by conducting look-ahead search from the current belief. During the search, the planner explicitly reasons about the interactions among traffic agents by applying \modelname to predict the future, and conditions the predictions on beliefs and road contexts. The planner then outputs optimal driving policies from the search tree and re-plans periodically.


In our experiments, we validate that \modelname is a realistic motion model by evaluating its prediction performance on real-world datasets containing both homogeneous and heterogeneous traffic, and report state-of-the-art results. Using \modelname, \simname generate complex, realistic mixed traffic in real-world urban environments. Our expert planner \algname also achieves superior driving performance in \simname, demonstrating safer, faster, and smoother driving compared with local collision avoidance and less-sophisticated planning methods. Open-source code of \simname, \modelname, and \algname are available online
\footnote{\href{https://github.com/AdaCompNUS/summit}{https://github.com/AdaCompNUS/summit}}\footnote{\href{https://github.com/AdaCompNUS/GAMMA}{https://github.com/AdaCompNUS/GAMMA}}\footnote{\href{https://github.com/AdaCompNUS/Context-POMDP}{https://github.com/AdaCompNUS/Context-POMDP}}.

\section{Related Work}

\begin{table}[!t]
\centering
\caption{Comparison between \simname and existing driving simulators.}
\begin{tabular}{ccccc }
\hline
\thead{Simulator} & \thead{Real\\-world \\
Maps} & \thead{Unregulated\\behaviors} & \thead{Dense\\Traffic} \tablefootnote{We only check-mark simulators explicitly featuring crowd behaviours.} &  \thead{Realistic\\Visuals \&\\Sensors} \\
\toprule
SimMobilityST \cite{SimMobilityST} & $\checkmark$ & $\times$ & $\checkmark$  & $\times$   \\
SUMO \cite{SUMO}&  $\checkmark$  &$\times$ & $\checkmark$ & $\times$ \\
TORCS \cite{TORCS}  & $\times$ & $\checkmark$ & $\times$ & $\checkmark$ \\
Apollo \cite{Apollo} & $\times$  & $\times$ & $\times$  & $\checkmark$\\
Sim4CV \cite{Sim4CV} & $\times$ & $\times$ & $\times$ & $\checkmark$ \\
GTAV \cite{GTAV}  & $\times$ & $\times$ & $\times$  & $\checkmark$ \\
CARLA \cite{CARLA}  & $\times$  & $\times$ & $\times$ & $\checkmark$ \\
AutonoViSim \cite{AutonoViSim} & $\times$ & $\checkmark$ & $\checkmark$ & $\checkmark$  \\
\thead{Force-based \\ simulator \cite{chao2019force}} & $\times$ & $\times$ & $\checkmark$ & $\checkmark$  \\
\simname (ours) & $\checkmark$  & $\checkmark$ & $\checkmark$ & $\checkmark$\\
\bottomrule
\end{tabular}
\label{tab::simulators}
\end{table}
\subsection{Existing Driving Simulators}
Driving simulators have been extensively applied to boost the development of autonomous driving systems.
Recent simulators (\tabref{tab::simulators}) have brought realistic visuals and sensors, but do not capture the complexities of urban environments and unregulated traffic behaviors.

Multi-car simulators like TORCS \cite{TORCS,Fluids,CoInCar-Sim} focus on interactions between multiple robot-vehicles. These simulators suit the study of complex interactions between agents, but can hardly scale up to crowded urban scenes.
CARLA \cite{CARLA}, Sim4CV \cite{Sim4CV}, and GTA \cite{GTAV} explicitly feature detailed physics modeling and realistic rendering for end-to-end learning. CARLA also provides a rich set of sensors such as cameras, Lidar, depth cameras, semantic segmentation, etc. However, these simulators rely on predefined maps, limiting the variety of environments. The simulated traffic also have relatively low density and simple rule-based behaviors. 
Another class of simulators \cite{SUMO,SimMobilityST,AutonoViSim,chao2019force} feature traffic simulation and control in urban environments. Among them, SUMO \cite{SUMO} and SimMobilityST \cite{SimMobilityST} support real-world maps but use simple rule-based behaviors, while another class \cite{AutonoViSim,chao2019force} apply more sophisticated motion models but are restricted to predefined maps. We aim to model the complexities in both urban maps and traffic behaviors in an automatic and unified framework.

\subsection{Traffic Motion Prediction and Crowd Simulation}
Existing traffic motion prediction and crowd simulation algorithms can be categorized into geometric and data-driven approaches. 

\subsubsection{Geometric Approaches}
Early models use social forces \cite{helbing1995social,lohner2010modeling,ferrer2013robot,pellegrini2009you} assuming traffic-agents are driven by attractive forces exerted by the destination and repulsive forces exerted by obstacles. These models can simulate large crowds efficiently, but the quality of interactions are often constrained by model simplicity. Otherwise, one need to manually engineer social force functions \cite{chao2019force} to achieve desirable quality in particular scenarios.

Velocity Obstacle (VO) \cite{fiorini1998motion} and Reciprocal Velocity Obstacle (RVO) \cite{van2008reciprocal,van2011reciprocal,snape2011hybrid} compute collision free motion by optimizing in the feasible velocity space. Variants such as GVO \cite{wilkie2009generalized}, NH-ORCA \cite{alonso2013optimal}, B-ORCA \cite{alonso2012reciprocal}, AVO \cite{van2011reciprocalAVO} explicitly handle non-holonomic traffic agents. However, these models require additional expertise such as explicitly-written kinematics or even dynamics equations. Some models also need additional manipulation, \eg, linearization, of the kinematics before incorporating them. Therefore, the above models do not generalize well to general heterogeneous kinematics of traffic agents.

Some previous work proposed different geometry representations for agents, such as ellipse \cite{best2016real}, capsule \cite{stuvel2017torso}, and medial axis transformation (CTMAT) \cite{ma2018efficient}. We found that polygons offers tight bounding to typical traffic agents with minimum trade-off of computational complexity. Some others also modeled behavioral types of crowd agents such as patience \cite{PORCA} and attention \cite{cheung2018efficient}. Here, we intend to capture the full set instead of a single aspect.

\subsubsection{Data-driven Approaches}

\begin{figure*}[!t]
\centering
\includegraphics[width=0.9\textwidth]{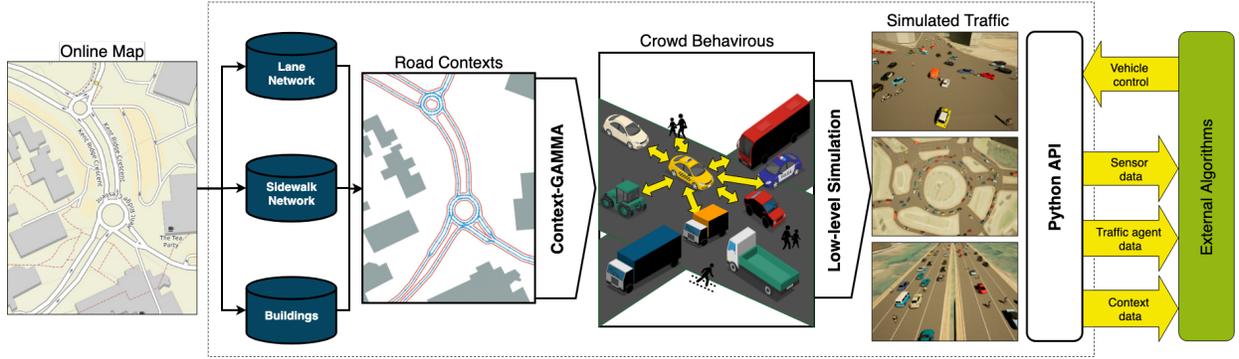}
\caption{An overview of \simname that simulates massive mixed traffic at any location in the world.}
\label{fig::overview}
\end{figure*}

Data-driven approaches learn to predict traffic motion in similar environments from calibrated data. Most data-driven methods are designed for homogeneous agents, mostly pedestrians. S-LSTM \cite{alahi2016social} models agents' interactions using a ``social pooling layer" to aggregate information from individual LSTMs. S-GAN \cite{gupta2018social} improves on S-LSTM by introducing a generative adversarial network architecture. SoPhie \cite{sadeghian2019sophie} leverages two attention modules to learn social interactions and physical constraints. SRLSTM \cite{zhang2019sr} leverages a state refinement module. 
S-Ways \cite{amirian2019social} uses Info-GAN\cite{chen2016infogan} for trajectory predictions. 
These methods have achieved high accuracy on pedestrian-only scenes. 
Some other work used similar recurrent neural networks for vehicle trajectory prediction \cite{altche2017lstm,deo2018convolutional,lee2017desire,kim2017probabilistic}. All above methods use identical architectures for independent traffic agents, thus can only handle homogeneous agents that look and behave similarly to each other.

In reality, traffic agents are often heterogeneous in kinematics and geometry. Recent models, like TraPHic \cite{chandra2018traphic}, TrafficPredict \cite{ma2018trafficpredict} and MATF \cite{zhao2019multi}, start to handle heterogeneous traffic. 
However, as deep neural networks do not explicitly understand underlying constraints and objectives, the above models can easily over-fit to characteristics of the training scenes. More generalizable models require much larger data sets, which are hard to acquire in practice. 

\subsection{Planning for Urban Driving}
While models described above can in principle be applied to perform driving in urban environments, they are usually not robust or safe enough due to the lack of long-term reasoning. Early long-term planning examples use multi-policy decision making \cite{galceran2015multipolicy,mehta2016autonomous} that roll-out a set of candidate policies into the future to choose an optimal one. These work focused on one-time decision-making in narrowly-defined scenarios \eg, lane merging, where candidate policies can be defined using domain knowledge. 
Other work applied POMDP planning \cite{bandyopadhyay2013intention,liu2015situation,sadigh2016information,chen2016pomdp,hubmann2018automated} to explicitly reason about exo-agents' uncertain intentions and types. However, early explorations have been restricted to few agents and simple motion models.

Recent work addresses driving among many pedestrians. The planner in \cite{PORCA} integrated a pedestrian motion model to performs POMDP planning to control the driving speed. Later, LeTS-Drive \cite{lets-drive} extended the work by integrating POMDP planning with learned heuristics. 
The \algname planner presented here can be viewed as extending \cite{PORCA} to the urban driving domain. Compared with driving among pedestrians, urban driving involves various new complexities: road contexts, heterogeneous agents, high-speed driving. These factors bring significantly challenges in both prediction and planning. 

\section{\simname Simulator}

We propose a new simulator, \simname, that focuses on simulating complex unregulated behaviours of dense urban traffic in real-world maps. It generates high-fidelity interactive data to facilitate the development, training, and testing of crowd-driving algorithms.
To construct real-world scenes, \simname fetches map data from the OpenStreetMap \cite{OSM}, and accordingly constructs two topological graphs - a lane network for vehicles, and a sidewalk network for pedestrians - as representations of road contexts. 
Then, \simname automatically generates a crowd and uses our motion model, \modelname, to simulate the motion of traffic agents. In the global-scope, \modelname takes road contexts as input to guide the traffic behaviors geometrically and topologically. In the local-scope, \modelname uses velocity-space optimization to generate sophisticated, unregulated crowd motion.
The low-level structure of \simname is based on CARLA, retaining its desirable features such as high-fidelity physics, realistic rendering, weather control, and rich sensors. 
We further developed an expert planner, \algname, using \simname to provide a reference for future crowd-driving algorithms.
\figref{fig::overview} provides an overview of \simname.

In this section, we introduce the representation of real-world maps and the Python API. Details of the \modelname motion model and the \algname planner will be presented in Section \ref{sec::gamma} and Section \ref{sec::planner}, respectively.
\subsection{Lane Network}
A lane network in \simname defines the connectivity of the road structure at the fidelity of individual lanes. The network consists of directed lane segments and connections between them. \simname relies on SUMO \cite{SUMO} to automatically convert OSM maps to lane networks. The extensive suite of network editing tools provided by SUMO can also be leveraged to improve and customize maps. The lane network interface allows users to locate traffic agents on the lane network and retrieve connected or neighbouring lane segments. The interface closely follows CARLA's waypoint interface, so that CARLA users can easily adapt to it.

\subsection{Sidewalk Network}
A sidewalk network in \simname defines the behaviors of pedestrians, which usually walk along road edges and occasionally cross roads. The network contains sidewalks near road edges, defined as poly-lines extracted from the road geometry, and cross-able roads, defined as connections between poly-lines. The sidewalk network interface allows users to locate pedestrians on the network, retrieve the current path, and retrieve the opposite sidewalk for potential road-crossing.

\subsection{Occupancy Map Interface}
We additionally provide an occupancy map interface to expose drive-able regions for the ego-vehicle. An occupancy map is the top-down projection of the road geometry that can be retrieved for any area. It can be used either for collision checking in control and planning algorithms or as bird-view input to neural networks.

\begin{figure*}[!t]
\centering
\begin{tabular}{ccc}
\includegraphics[height=1.7in]{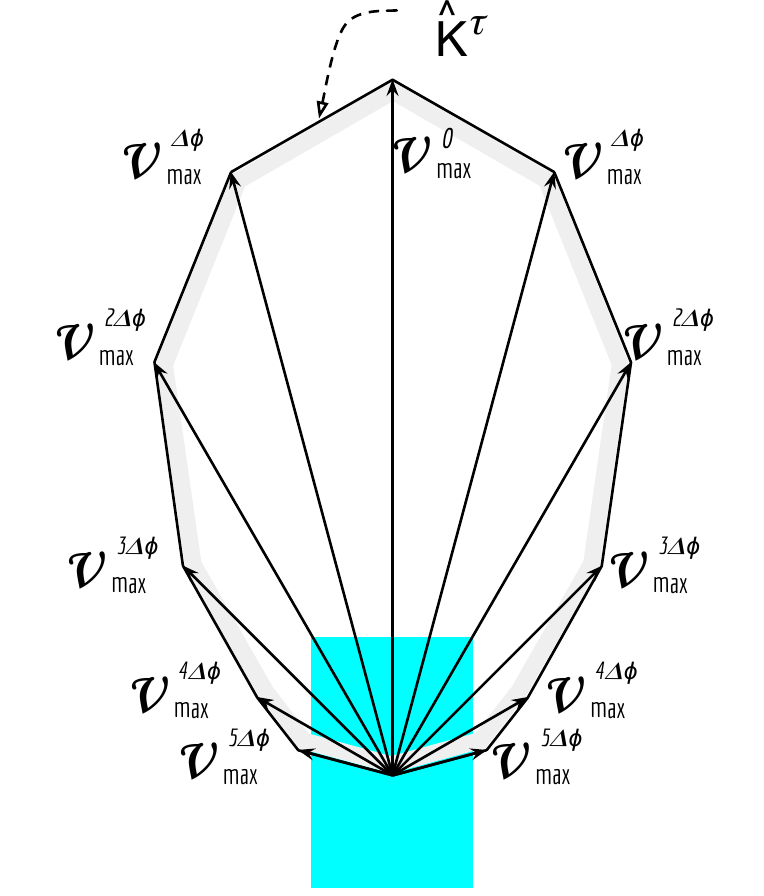}&
\includegraphics[height=1.7in]{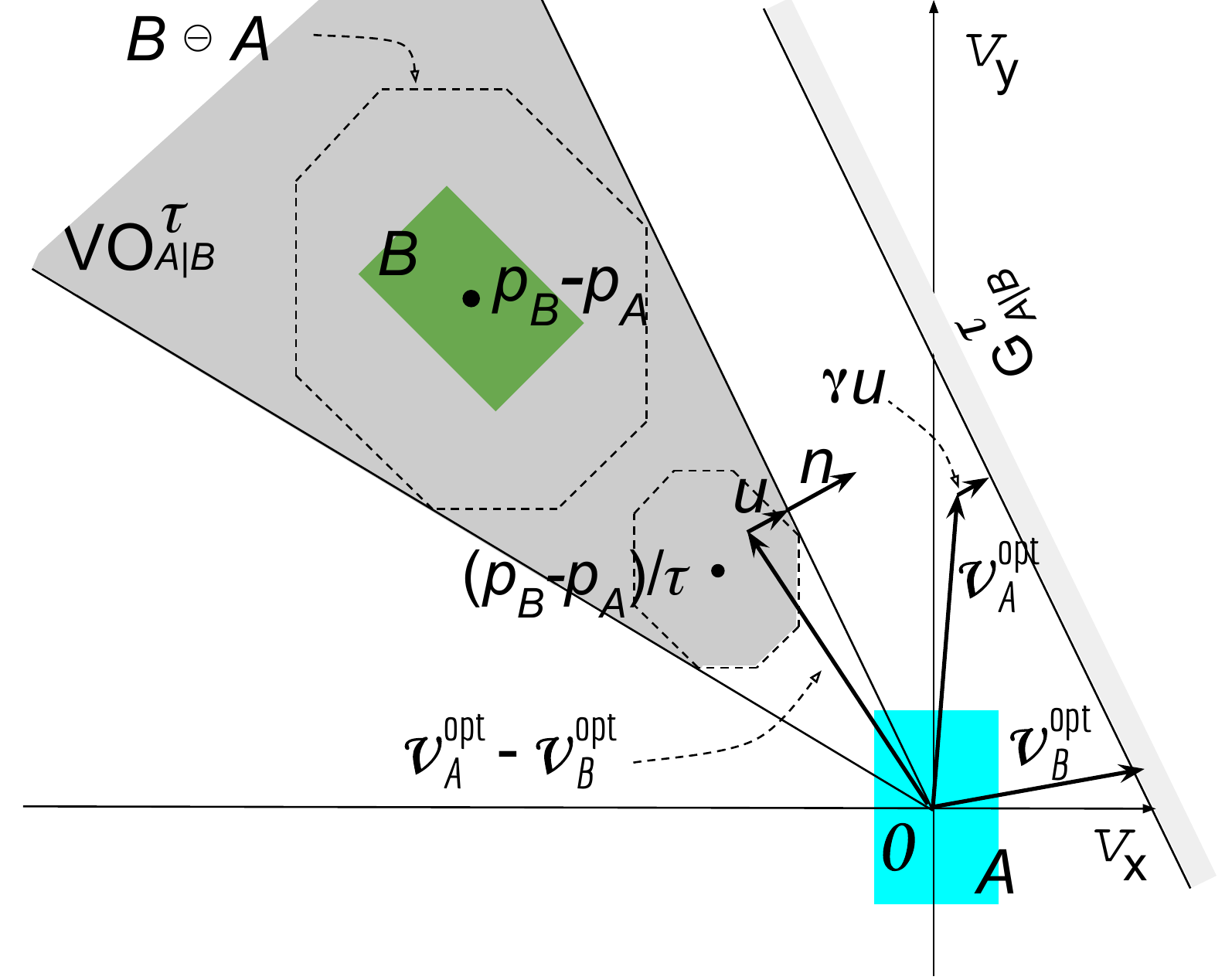} &
\hspace*{0.3cm}
\includegraphics[height=1.7in]{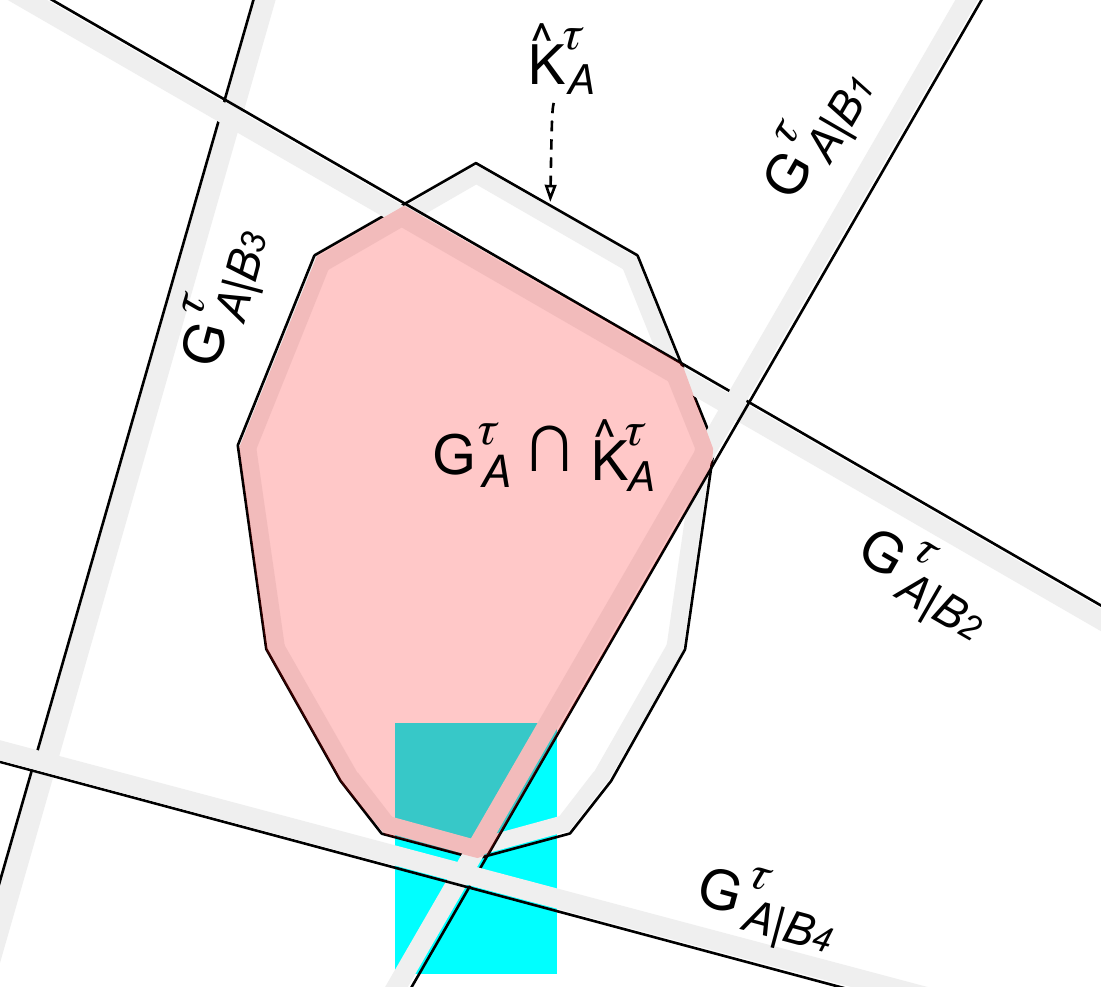} \\
      (\subfig a) & (\subfig b) & (\subfig c)
\end{tabular}  
 \caption{(\subfig a) The estimated kinematically trackable velocity set, $\hat{\kinset}^\timewin$, is estimated using the convex hull of $\{ v^\phi_{\mathrm{max}} \mid \phi \in \Phi\}$. (\subfig b) $\vo_{A|B}^\timewin$ (gray) and $\geoset_{A|B}^\timewin$ (half plane). The velocity obstacle of agent $A$ (blue) induced by agent $B$ (green) for time $\timewin$, is a truncated cone with its apex at the origin (in velocity space) and its legs tangent to the polygon $B \ominus A$. $\geoset_{A|B}^\timewin$ is a half plane divided by the line perpendicular to the vector $\velu$ through the point $\vel_A^{\mathrm{opt}}+\responsibility\velu$, where $\velu$ is the vector from $\vel_A^{\mathrm{opt}} - \vel_B^{\mathrm{opt}}$ to the closest point on the boundary of $\vo_{A|B}^\timewin$. (\subfig c) An example of the feasible velocity space of agent $A$. The feasible space (red), is the intersection of $\hat{\kinset}^\timewin_A$ and the geometry constraints induced by four other agents: $\geoset_{A|B_1}^\timewin$,$\geoset_{A|B_2}^\timewin$,$\geoset_{A|B_3}^\timewin$, and $\geoset_{A|B_4}^\timewin$.}\label{fig:vo-g-k-f}
   \end{figure*}
   
\subsection{Landmarks}
\simname also makes use of landmark data in OSM maps such as buildings and forests to provide structurally rich and realistic visuals. We additionally support randomization of the landmark textures to generate more versatile visual inputs and enable techniques such as domain randomization \cite{domain}.

\subsection{Python API}
The Python API of \simname extends that of CARLA with additional functionalities. The API allows users to fetch new real-world maps, spawn random crowds, and configure the parameters of \modelname-controlled agents. It exposes to external algorithms not only sensor data and agent states, but also road contexts like lane networks, sidewalk networks, and occupancy maps. 
Users can send back vehicle controls - steering, acceleration, braking, reversing, \etc - to drive a vehicle and interact with the environment. 
This extended API is designed to enable a wide range of applications such as perception, sensor-based control, model-based reasoning, and end-to-end learning.

\section{The \modelname Motion Model} \label{sec::gamma}
The \modelname motion model is the core of \simname that generates complex crowd bahaviours.
We formulate the motion of traffic agents as constrained geometric optimization in velocity space and build our model based on this formulation. Particularly, \modelname assumes each traffic agent optimizes its velocity based on the navigation goal, while being constrained by the following constraints:
\begin{itemize}
    \item kinematic constraints: car-like agents conduct non-holonomic motion, \etc.
    \item geometric constraints: agents try to avoid collision with nearby agents;
    \item context constraints: agents tend to conform with road structures, \eg, avoid going off-road or driving in the opposite direction \etc;
\end{itemize}
A velocity $\vel$ for a traffic agent $A$ is called \emph{kinematically trackable} if $A$ can track $\vel$ with its low-level controller below a predefined maximum tracking error $\maxbounderror$ for $\timewin$ time.
It is further called \emph{geometrically feasible} if it does not lead to collisions with any other nearby agents by taking $\vel$ for $\timewin$ time. 
Finally, $\vel$ is called \emph{contextually conformed} if it does not violate the hard contextual constraints for $\timewin$ time.

Let velocity sets $\kinset_A^\tau$, $\geoset_A^\tau$ and $\conset_A^\tau$ represent the kinematically trackable set, geometrically feasible set and the contextually conformed set, respectively. 
\modelname selects for $A$ a new velocity from the intersection:
\begin{equation}\label{eq:gamma-obj-fun}
v_A^\mathrm{new} = \argmin_{v \in \kinset_{A}^{\tau} \cap \geoset_{A}^{\tau} \cap \conset_{A}^{\tau}} \norm{v - v_A^\mathrm{pref}}
\end{equation}
where $v_A^\mathrm{pref}$ is $A$'s preferred velocity computed from its intention.
The actual motion of $A$ is generated by tracking $\vel_A^\mathrm{new}$ using $A$'s low-level controller. 

The following sections introduce the construction of $\kinset_A^\tau$, $\geoset_A^\tau$ and $\conset_A^\tau$, respectively.

\subsection{Kinematics Constraints} \label{sec:kinematics}

Motions of real-world traffic agents are constrained by their kinematics: pedestrians are holonomic; vehicles are non-holonomic and cannot move sidewise; trailers are not actuated and can only follow the head truck.
\modelname only allows kinematically-trackable velocities w.r.t. a maximum error $\maxbounderror$ and a time window $\tau$.
The challenge here is to model \emph{heterogeneous} kinematics including holonomic (pedestrians), car-like (cars, bicycles), trailer-like (trucks) ones, \etc.

Concretely, for a given traffic agent $A$, we aim to find the \emph{kinematically trackable velocity set}, $\kinset^\timewin_A$, \emph{s.t.},
\begin{equation}\label{eq:kinematic}
\kinset^\timewin_A = \{ \vel \mid \norm{\vel t - \kinfun_A(\vel,t)} \leq \maxbounderror , \forall t \in [0, \timewin] \},
\end{equation}
where $\kinfun_A(\vel,t)$ is the position of $A$ at time $t$ if it tracks $\vel$ with its low-level controller. As $\kinfun_A(\vel,t)$ varies for different kinematics and controllers, it is often intractable to compute its analytic form. We propose to estimate $\kinset^\timewin_A$ numerically. 

Given a type of agent, we estimate its $\kinset^\timewin$ using the following procedure.
The velocity set $V$ is first discretized along its two dimensions $(s,\phi)$, where $s$ is the speed and $\phi$ is the deviation angle from traffic agent's heading direction. We consider discrete sets of values in each dimension: $\hat{S} = [0:\Delta s:s_\mathrm{max}]$ and $\hat{\Phi} = [0:\Delta \phi:\phi_\mathrm{\max}]$.
The Cartesian product of these sets form a discretized velocity set $\hat{V} = \{ v=(s_v,\phi_v) \mid s_v \in \hat{S}, \phi_v \in \hat{\Phi}\}$.
Along each deviation angle $\phi \in \hat{\Phi}$, we run the controller for a duration of $\timewin$ to track each speed $s$ and record the maximum tracking error $\varepsilon_{(s,\phi)}$. We then pick the maximum speed along direction $\phi$ that satisfies $\varepsilon_{(s,\phi)}<\maxbounderror$, and register it as a boundary point of $\kinset^\timewin$, denoted as $v^\phi_{\mathrm{max}}$. After enumerating all deviation angles, we can approximate $\kinset^\timewin$ by the convex hull of boundary velocities (\figref{fig:vo-g-k-f}\subfig{a}):
\begin{equation}
\approxkinset^\timewin = \mathrm{ConvexHull}(\{ \vel^\phi_{\mathrm{max}} \mid \phi \in \Phi\})
\end{equation}

Construction of $\kinset^\timewin$ is conducted offline for each representative type of agents: pedestrian, bicycle, motorbike, car, van, bus, gyro-scooter, trucks, \etc. Although approximating $\kinset^\timewin$ brings approximation error, we found that \modelname is robust to this error empirically.

\subsection{Geometry Constraints}\label{sec:geometry}
To model collision avoidance, we first determine for each agent $A$ its velocity obstacle $\vo_{A}^\timewin$ w.r.t. other agents, then derive an optimal collision avoidance velocity set $\geoset{}_{A|B}^{\timewin}$ from $\vo_{A}^\timewin$. Our definition of $\geoset{}_{A|B}^{\timewin}$ extends those in \cite{van2011reciprocal} by replacing disc-shaped agents by polygons, which offer tighter fits to typical traffic agents.

Consider two agents $A$ and $B$ at position $\pose_A$ and $\pose_B$, respectively. The \emph{velocity obstacle} $\vo_{A|B}^\timewin$ is defined as the set of \emph{relative} velocities of $A$ w.r.t. $B$ that result in collisions before time $\timewin$. Formally,
\begin{equation}
\label{eq:vo-general}
\vo_{A|B}^\timewin = \{\vel \mid \exists t \in [0,\timewin], t \cdot \vel \in (B \ominus A) \},
\end{equation}
where $B \ominus A$ represents the \textit{Minkowski difference}, which inflates the polygonal geometry of $B$ by that of $A$ so that $A$ can be treated as a single point. \figref{fig:vo-g-k-f}\subfig{b} visualizes a velocity obstacle constructed with two polygon-shaped agents.

We construct $\geoset{}_{A|B}^{\timewin}$ w.r.t. the current velocities of the agents.
Suppose $A$ and $B$ collide with each other before time $\timewin$ by taking their current velocities $\vel_A^\mathrm{cur}$ and $\vel_B^\mathrm{cur}$.
To avoid collisions with the least cooperative effort, \modelname finds a relative velocity closest to $\vel_A^\mathrm{cur} - \vel_B^\mathrm{cur}$ from the boundary of $\vo_{A|B}^\timewin$. Let $\velu$ be the vector from $\vel_A^\mathrm{cur} - \vel_B^\mathrm{cur}$ to this optimal relative velocity: 
\begin{equation} \label{eq:velu}
\velu = (\argmin_{\vel \in \partial \vo_{A|B}^\timewin} \norm{\vel-(\vel_A^\mathrm{cur} - \vel_B^\mathrm{cur}) }) - (\vel_A^\mathrm{cur} - \vel_B^\mathrm{cur}).
\end{equation}
Then $\velu$ is the smallest change on the relative velocity to avoid the collision within $\timewin$ time. 
\modelname lets $A$ take $\responsibility{} \in [0,1]$ of the responsibility for collision avoidance, \ie, adapt its velocity by $\responsibility{} \cdot \velu$. Then $\geoset_{A|B}^\timewin$ is constructed as a half plane:
\begin{equation}
\label{eq:geo_A_B}
\geoset_{A|B}^\timewin = \{ \vel \mid (\vel - (\vel_A^\mathrm{cur}+\responsibility{} \cdot \velu))\cdot \veln \geq 0 \},
\end{equation}
where $\veln$ is the outward normal at point $(\vel_A^\mathrm{cur} - \vel_B^\mathrm{cur}) + \velu$. 
\figref{fig:vo-g-k-f}\subfig{b} visualizes $\geoset_{A|B}^\timewin$. 
Note that the responsibility \responsibility{} is an inner state of human that can be inferred from interaction history (\secref{sec:bayesian-inference}). 

Finally, the geometrically feasible velosity space of $A$ is constructed by considering all nearby agents:
\begin{equation}
\label{eq:geo_A}
\geoset{}_{A}^{\timewin} = \bigcap_{B \in \attfun{}(A)} \geoset{}_{A|B}^{\timewin}
\end{equation}
We have further assumed that traffic agents only have limited attention capabilities, and can only attend to agents within its attentive range denoted as $\attfun{}(A)$.

We use two half circles to model agent's attention: one in front of the vehicle with radius $r_\mathrm{front}$, and one at the back with radius $r_\mathrm{rear}$. We set $r_\mathrm{rear} \leq r_\mathrm{front}$, such that agents give more attention to others in the front. The actual values of $r_\mathrm{front}$ and $r_\mathrm{rear}$ can also be inferred from history (\secref{sec:bayesian-inference}). 

Since $\geoset_{A|B}^\timewin$ are half-planes in the velocity space, their intersection $\geoset{}_{A}^{\timewin}$ is convex. So does its intersection with $\kinset_A^\timewin$.
\figref{fig:vo-g-k-f}\subfig{b} visualizes an example of $\geoset{}_{A}^{\timewin}$ induced by four other agents, and \figref{fig:vo-g-k-f}\subfig{c} shows it overlaid on $\kinset_A^\timewin$.

\subsection{Contextual Constraints}\label{sec:context}

\modelname can also model road structures and traffic rules.
This is achieved by injecting an additional set of \emph{contextual constraints} in agents' velocity space, denoted as $\conset_A$ for agent $A$. 

Take driving direction for example.
To prevent car $A$ from driving to the opposite lane within a time window $\tau_1$, the lateral speed of the car should be constrained under $d_1/\tau_1$, where $d_1$ is the distance from the car to the opposite lane. This constraint forms a half-plane in the velocity space, $\conset_{A|C_{\mathrm{opp}}}^{\tau_1}$, defined by a separation line parallel to the opposite lane with an offset of $d_1/\tau_1$ from the car (the origin of the velocity space). Any velocity in $\conset_{A|C_{\mathrm{opp}}}^{\tau_1}$ would be feasible.


\subsection{Optimization Objective}
To predict the motion of $A$, \modelname computes a velocity closest to its \textit{preferred velocity} $\vel_A^\mathrm{pref}$ according to \eqref{eq:gamma-obj-fun}.
The preferred velocities are derived from traffic agents' intentions, which are defined differently for cases where contextual information exists, as in simulation, and where contexts do not exist, as in existing data sets. 

When road contexts are available, agent intentions are defined as the intended paths extracted from the lane and sidewalk networks. Given a path, we first extract a look-ahead way-point, then point $\vel^{\mathrm{pref}}$ towards the way point.
When road contexts are not available, like in most existing real-world datasets, we adopted a context-free intention representation characterizing whether an agent wants to keep moving at current velocity or at current acceleration. 
We then apply Bayesian inference (\secref{sec:bayesian-inference}) to infer the actual, hidden intentions based on agents' history positions. 

The objective function in \eqref{eq:gamma-obj-fun} is quadratic. Moreover, since $\geoset{}_{A}^{\timewin}$, $\kinset{}^\timewin_A$ and $C_{A}^{\timewin}$ are all convex by construction, the optimization problem defined in \eqref{eq:gamma-obj-fun} can be efficiently solved in linear time. We thus efficiently compute new tracking velocities of all agents at each time step. The actual motion of agents are generated by their low-level controllers.

\section{Context-aware POMDP Planning} \label{sec::planner}
\begin{figure}
    \centering
    \includegraphics[width=0.95\columnwidth]{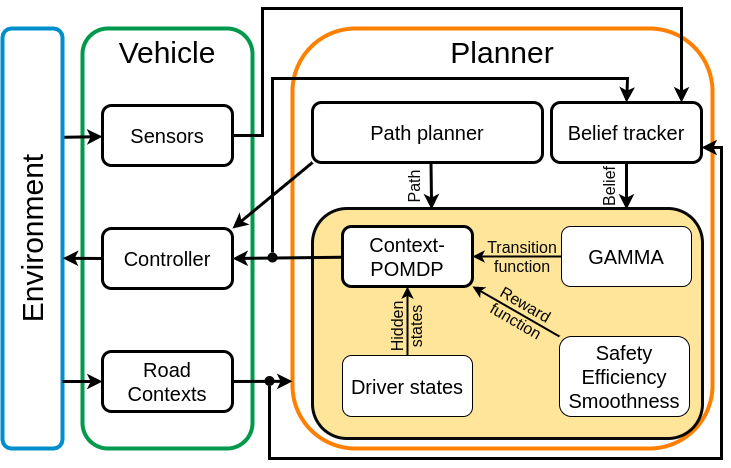}
    \caption{Architecture of our crowd-driving planner that integrates GAMMA as the transition model for exo-agents. The core of the planner is the \algname model that is conditioned on road contexts and uncertain states of human drivers.}
    \label{fig:pomdp_planner}
\end{figure}
We then present the \algname planner which is developed in \simname. 
Planning for driving in an unregulated dense traffic is extremely challenging. The robot has to be smart enough to make efficient progress, instead of being "frozen" and stuck in the crowd. In the meantime, a highly dynamic and interactive crowd makes the task safety critical. Mistakes in planning can lead to severe or even fatal accidents.

The key of success is to explicitly model the uncertainty on \internalstates{} and interactions among traffic agents. 
We formulate crowd-driving as a POMDP problem, treating \internalstates{} as hidden variables, using \modelname to predict future interactions, and conditioning the problem on road contexts to reduce the complexity. The expert planner solves this POMDP efficiently using a state-of-the art belief tree search algorithm. The planner is thus referred to as \algname.

\figref{fig:pomdp_planner} shows the overall architecture of the crowd-driving system. 
At each time step, the system receives an observation from \simname and uses it to update its belief over \internalstates{}. Asynchronously, a route for the ego-vehicle is determined using a path selector, \eg, a simple A* search on the lane network. The updated belief and the route are then passed to \algname to solve for the optimal acceleration along the driving path. 

\algname treats two \internalstates{}, intentions and attentiveness, as hidden variables. It uses \modelname as the transition model to predict the interactive future and optimizes ego-vehicle's policy accordingly. The objective is to maximize the safety, efficiency and smoothness of driving.

\subsection{\algname}
In this section, we present our POMDP model for contextual crowd-driving.

\subsubsection{State and Observation Modelling} \label{sec:hidden}

A state in \algname is composed of the following observable states:
\begin{itemize}
  \item State of the ego-vehicle, $s_c=(x,y,\phi,\mu)$, including the position $(x,y)$, heading direction $\phi$, and the intended driving path $\mu$;
  \item States of exo-agents, $\{s_i=(x,y,\vec{v})\}_{i\in I_{exo}}$, including the position $(x,y)$ and the current velocity $\vec{v}$. $I_{exo}$ defines the set of indices of exo-agents; 
\end{itemize}
and two hidden variables for each exo-agent representing critical \internalstates{}, $\{\theta_i=(\mu_i, t_i)\}_{i\in I_{exo}}$:
\begin{itemize}
  \item The \textit{intention} $\mu_i$ of agent $i$: Let $U_i$ be the set of path candidates for the traffic agent, which are extracted from the road contexts such as the lane network and the sidewalk network. This agent may take any of the path candidates $\mu_i \in U_i$ as its actual intention. 
  \item The \textit{type} $t_i$ of agent $i$: An agent can be either \textit{distracted}, thus not interacting with the ego-vehicle, or be \textit{attentive}, thus cooperatively avoid collision with the ego-vehicle (modelled using \modelname).
\end{itemize}

We assume that the ego-vehicle can observe discretized values of the observable states. However, inner states of exo-agents can only be inferred and modelled with beliefs. At the beginning of an episode, the initial belief over \internalstates{} is set to be uniform for all observed agents. The belief is updated at every time step using Bayesian filtering (see Section \ref{sec:bayesian-inference}) and serves as the inputs to online planning.

\subsubsection{Action Modelling}
The action space of the ego-vehicle consists of its steering angle and acceleration. Given the well-known exponential complexity of POMDP planning \cite{Kaelbling_1998}, \algname decouples the action space of the ego-vehicle to keep the branching factor of the planning problem within a tractable range. Concretely, we restrict the POMDP to compute the acceleration along the intended path, while the steering angle is generated separately using a rule-based lane-changing decision-maker and a pure-pursuit algorithm. The action space contains three possible accelerations for each time step: $\{ACC, MAINTAIN, DEC\}$. The acceleration value for $ACC$ and $DEC$ is $ \SI{3}{\meter \per \second^2}$ and $\SI{-3}{\meter \per \second^2}$, respectively. The maximum speed of the ego-vehicle is $\SI{6}{\Ms}$.
\subsubsection{Transition Modelling \label{sec::transition}}
\algname predicts traffic agents' motion using the following set of models. Distracted traffic agents are assumed to track their intended path with the current speed. Attentive traffic agents also tend to follow the sampled path, but use \modelname to generate the collision avoidance behaviours. The motion of all agents, including the ego-vehicle, are constrained by their kinematics, \eg, pedestrians are simulated using holonomic motion and car-like vehicles are simulated using bicycle models. To model stochastic transitions of exo-agents, their displacements are perturbed by Gaussian noises.
\subsubsection{Reward Modelling}
The reward function in \algname takes in to account safety, efficiency, and smoothness of driving. It assigns large penalties $R_\mathrm{col}=-3000\times(v^2 + 0.5)$ when the ego-vehicle collides with any exo-agent at speed $v$, uses a speed cost $R_\mathrm{speed} = \frac{v - v_\mathrm{max}}{v_\mathrm{max}}$ to penalize driving at low speed, and finally, penalizes frequent deceleration using a penalty of $R_\mathrm{acc} = -0.1$ for each deceleration. 



\subsection{Belief tracking for Human Inner States}
\label{sec:bayesian-inference}
Intentions and types of agents are inner states of human that can change the crowd behaviours but are not observable in practice. We can only infer a distribution, or belief, over their possible values from agents' history. This belief is tracked using Bayesian filtering.


Concretely, belief tracking over \internalstates{} is implemented as a factored histogram filter. A belief is represented as a discrete set of probabilities over all possible values of inner states for all agents. 
We assume that the inner states of a traffic agent is static during interactions. 
At each time step, 
for a particular hypothesis of inner state values $\theta$ of a particular agent, we assume that the \textit{expected} next position of the agent is determined by \modelname and $\theta$: 
\begin{equation}
\label{eq:dynamic-model}
\bar{\pose}_{t} = \pose_{t}^{\mathrm{\modelname}}(\theta, \pose_{t-1}),
\end{equation}
where $\pose_{t-1}$ is the history position at time $t-1$.

After observing the \textit{actual} next position $\pose_{t}$ of the agent, the observation likelihood for hypothesis $\theta$ is computed as:
\begin{equation}
\label{eq:motion-dynamic-model}
p(\pose_{t} \mid \theta) = f(\norm{\pose_{t} - \pose_{t}^{\mathrm{\modelname}}} \mid 0, \sigma^2),
\end{equation}
where $f$ is the probability density function of normal distributions. We assume the transition noise is Gaussian and has a predefined variance $\sigma^2$.

Now, we update the posterior distribution over $\theta$ for the agent with the observation likelihood $p(\pose_{t} \mid \theta)$, using the Bayes' rule:
\begin{equation}
\label{eq:bayes-rule-his}
p_{t}(\theta) = \eta \cdot p(\pose_{t} \mid \theta) \cdot p_{t-1}(\theta),
\end{equation}
where $\eta$ is a normalization constant. 

This process is repeated for all agents and all values of inner states to output the updated joint belief, which is then passed to the planner. This histogram filter is also used for predicting the motion of real-world agents in our experiments (\secref{sec:exp-of-gamma}).

\subsection{Solving the \algname}
The \algname planner solves the POMDP model to produce real-time decisions for the ego-vehicle. This is achieved using a state-of-the-art belief tree search algorithm, HyP-DESPOT. HyP-DESPOT conditions belief tree search on a small set of sampled scenarios to reduce the computational complexity while performing near-optimal planning. It further uses parallel tree search to scale-up to complex tasks. Here, we only introduce the high-level structure of the planner and refer readers to \cite{Hyp-despot} for details on HyP-DESPOT.

At each time step, the planner builds a sparse belief tree with the root being the current belief and each node denoting a reachable future belief. The tree branches recursively with future robot actions (lane decisions and accelerations) and observations (agent states). After searching a sufficiently complete tree, the planner outputs the optimal policy (also a tree of future beliefs) under the current belief. It then executes the first action in the policy on the ego-vehicle, receive new observations through the Python API of \simname, and update the belief over \internalstates{}. This planning process is repeated for every time step at a rate of 3 HZ.

\section{Results}
We want to answer the following questions in the experiments:
\begin{itemize}
    \item Is \modelname a realistic motion model? Does it generate accurate predictions for real-world traffic?
    \item Can \simname simulate complex scenes? Does the simulated crowd reflect the objective of human participants?
    \item Can \algname drive a vehicle safely and efficiently in dense unregulated urban traffic? Is it better than other simpler solutions?
\end{itemize}
We provide both qualitative and quantitative results to answer these questions.


\subsection{Prediction Performance on Real-world Datasets}

\label{sec:exp-of-gamma}

In this section, we aim to validate that \modelname is a realistic motion model. To do so, we evaluate the prediction accuracy of \modelname on four real-world datasets: ETH\cite{pellegrini2010improving}, UCY\cite{leal2014learning}, UTOWN, and CROSS. 

To predict the motion of real-world agents, we integrate \modelname with the histogram filter described in \secref{sec:bayesian-inference} to infer the distribution over \internalstates{} - intention, attention, and responsibility - from agent's history and condition preditions on the distribtution.
We first evaluate \modelname on \emph{homogeneous} datasets with only pedestrians (ETH and UCY) in \secref{sec:heterogeneous-agents-datasets}, then on \emph{heterogeneous} datasets with various types of agents (UTOWN and CROSS) in \secref{sec:pedestrian-datasets}. 
Finally, we conduct an ablation study in \secref{sec:ablation-study}.


We compare \modelname with both geometry-based \cite{PORCA,yamaguchi2011you} and data-driven models 
\cite{alahi2016social,zhang2019sr,ma2018trafficpredict,gupta2018social,sadeghian2019sophie,zhao2019multi,amirian2019social}.
Following prior work \cite{lee2017desire,gupta2018social}, we report two error metrics:
\begin{itemize}
   \item{Average Displacement Error (\textbf{ADE}).} The average Euclidean distance (in meter) between the predicted position and the ground-truth position over all the time frames for a prediction duration $t_\text{pred} = 12$ $(4.8s)$.
   \item{Final Displacement Error (\textbf{FDE}).} The Euclidean distance (in meter) between the predicted position and the ground-truth position at the final time frame.
\end{itemize}

\subsubsection{Evaluation on Homogeneous Datasets}
\label{sec:pedestrian-datasets}

\begin{table*}[t]
\centering
\caption{Performance comparison across homogeneous benchmark datasets, ETH (ETH and HOTEL) and UCY (UNIV, ZARA1, and ZARA2). The ADE and FDE values are separated by slash. Their average values (AVG) are also reported.}\label{tab:learning-based-approaches} 
{\scriptsize
\begin{tabular}{cccccccccccccc}
\toprule
\multicolumn{8}{c}{\textbf{Deterministic Models}}  & \multicolumn{5}{c}{\textbf{Stochastic Models}}  \\ 
Dataset&\hspace{-5pt} LR      &\hspace{-5pt} S-LSTM     &\hspace{-5pt} SRLSTM               &\hspace{-5pt}TrafficPredict   &\hspace{-5pt}S-Force   &\hspace{-5pt}PORCA       &\hspace{-5pt} GAMMA                        &\hspace{-5pt} SGAN      &\hspace{-5pt} SGAN-ave  &\hspace{-5pt}SoPhie      &\hspace{-5pt}MATF                          &\hspace{-5pt}S-Ways                        & \hspace{-5pt}GAMMA-s  \\ 
\cmidrule(lr){0-7}\cmidrule(lr){9-14}
ETH   &\hspace{-5pt} 1.33/2.94 &\hspace{-5pt} 1.09/2.35  &\hspace{-5pt}0.63/1.25                    &\hspace{-5pt}5.46/9.73 &\hspace{-5pt}0.67/1.52 &\hspace{-5pt}0.52/1.09  &\hspace{-5pt} \textbf{0.51}/\textbf{1.08}  &\hspace{-5pt} 0.81/1.52 &\hspace{-5pt} 0.96/1.87 &\hspace{-5pt} 0.70/1.43  &\hspace{-5pt} 1.01/1.75                    &\hspace{-5pt} \textbf{0.39}/\textbf{0.64}  & \hspace{-5pt}0.43/0.91     \\ 
HOTEL &\hspace{-5pt} 0.39/0.72 &\hspace{-5pt} 0.79/1.76 &\hspace{-5pt}0.37/0.74                     &\hspace{-5pt}2.55/3.57 &\hspace{-5pt}0.52/1.03 &\hspace{-5pt}0.29/0.60  &\hspace{-5pt} \textbf{0.28}/\textbf{0.59}  &\hspace{-5pt} 0.72/1.61 &\hspace{-5pt} 0.61/1.31 &\hspace{-5pt} 0.76/1.67  &\hspace{-5pt} 0.43/0.80                    &\hspace{-5pt} 0.39/0.66                    & \hspace{-5pt}\textbf{0.23}/\textbf{0.48} \\   
UNIV  &\hspace{-5pt} 0.82/1.59 &\hspace{-5pt} 0.67/1.40  &\hspace{-5pt}0.51/1.10  &\hspace{-5pt}4.32/8.00 &\hspace{-5pt}0.74/1.12 &\hspace{-5pt}0.47/1.09  &\hspace{-5pt} \textbf{0.44}/\textbf{1.06}                    &\hspace{-5pt} 0.60/1.26 &\hspace{-5pt} 0.57/1.22 &\hspace{-5pt} 0.54/1.24  &\hspace{-5pt} 0.44/\textbf{0.91}                    &\hspace{-5pt} 0.55/1.31                    & \hspace{-5pt}\textbf{0.43}/1.03  \\    
ZARA1 &\hspace{-5pt} 0.62/1.21 &\hspace{-5pt} 0.47/1.00  &\hspace{-5pt}0.41/0.90  &\hspace{-5pt}3.76/7.20 &\hspace{-5pt}0.40/0.60 &\hspace{-5pt}0.37/0.87  &\hspace{-5pt} \textbf{0.36}/\textbf{0.86}                    &\hspace{-5pt} 0.34/0.69 &\hspace{-5pt} 0.45/0.98 &\hspace{-5pt} 0.30/0.63  &\hspace{-5pt} \textbf{0.26}/\textbf{0.45}  &\hspace{-5pt} 0.44/0.64                    & \hspace{-5pt}0.32/0.75  \\ 
ZARA2 &\hspace{-5pt} 0.77/1.48 &\hspace{-5pt} 0.56/1.17  &\hspace{-5pt}0.32/0.70                    &\hspace{-5pt}3.31/6.37 &\hspace{-5pt}0.40/0.68 &\hspace{-5pt}0.30/0.70  &\hspace{-5pt} \textbf{0.28}/\textbf{0.68}  &\hspace{-5pt} 0.42/0.84 &\hspace{-5pt} 0.39/0.86 &\hspace{-5pt} 0.38/0.78  &\hspace{-5pt} 0.26/\textbf{0.57}           &\hspace{-5pt} 0.51/0.92                    & \hspace{-5pt}\textbf{0.24}/0.59 \\
\cmidrule(lr){0-13}
AVG   &\hspace{-5pt} 0.79/1.59 &\hspace{-5pt} 0.72/1.54  &\hspace{-5pt}0.45/0.94                    &\hspace{-5pt}3.88/6.97 &\hspace{-5pt}0.55/0.99 &\hspace{-5pt}0.39/0.87  &\hspace{-5pt} \textbf{0.37}/\textbf{0.85}  &\hspace{-5pt} 0.58/1.18 &\hspace{-5pt} 0.60/1.25 &\hspace{-5pt} 0.54/1.15  &\hspace{-5pt} 0.48/0.90                    &\hspace{-5pt} 0.46/0.83                    & \hspace{-5pt}\textbf{0.33}/\textbf{0.75}  \\ 
\bottomrule  
\end{tabular}
}
\end{table*}

Most existing models are designed for pedestrians. We thus compare \modelname with these models on pedestrian-only datasets, ETH and UCY. 
We consider two versions of \modelname: a deterministic version that uses the most-likely inner state to generate predictions, and a stochastic version sampling 20 hypotheses from the inferred distribution and choose the maximum prediction accuracy. The purpose of adding the stochastic version is to conduct a fair comparison between the stochastic 20-max results reported in SGAN, MATF, SoPhie, and S-Ways. We refer to the stochastic version as \modelname-s.
Choosing the best sample, however, is meaningless for autonomous driving since it requires the hindsight knowledge of the ground truth. 
One way to make those stochastic approaches meaningful is to compute an average sample and use the average sample as the prediction instead of using the one closest to the ground truth. Following this idea, we modified SGAN and get an additional baseline SGAN-ave.

\tabref{tab:learning-based-approaches} shows the ADE and FDE of \modelname and the baselines. \modelname outperforms all existing models, including deterministic models SRLSTM and PORCA with deterministic prediction, and stochastic models MATF and S-Ways with stochastic measures. 
The performance gain over PORCA is not significant here because the datasets only contains pedestrians that conduct holonomic motion and fit well with disk-like shapes. We will show in \secref{sec:heterogeneous-agents-datasets} that \modelname outperforms PORCA significantly on heterogeneous datasets.

One interesting finding on data-driven approaches is that, the simple linear model significantly outperforms most deep learning approaches including LSTM, S-LSTM, SGAN and SoPhie in the HOTEL dataset where the scene is less crowded and trajectories are mostly straight-lines. 
This indicates that data-driven models over-fit to data in complex scenes, and fail to handle the simplest cases.
In contrast, \modelname performs well in both simple and complex scenes.

\subsubsection{Evaluation on Heterogeneous Datasets}
\label{sec:heterogeneous-agents-datasets}


\begin{table}[t]
\centering
\caption{Performance comparison across heterogeneous datasets CROSS and UTOWN. 
ADE/FDE, and their mean (AVG) are reported.}\label{tab:traditional-approaches}
{\scriptsize
\begin{tabular}{ccccccc}
\toprule
Dataset &\hspace{-9pt}TrafficPredict &\hspace{-9pt} SRLSTM      &\hspace{-9pt} SGAN-ave    &\hspace{-9pt} SGAN       &\hspace{-9pt} PORCA             &\hspace{-9pt} GAMMA         \\ 
\cmidrule(lr){0-6}
CROSS   &\hspace{-9pt} 6.49/11.56    &\hspace{-9pt} 1.36/3.17   &\hspace{-9pt} 1.16/2.66   &\hspace{-9pt} 0.93/2.18  &\hspace{-9pt} 0.89/2.10          &\hspace{-9pt} \textbf{0.64}/\textbf{1.77} \\ 
UTOWN   &\hspace{-9pt} 2.82/4.24     &\hspace{-9pt} 0.41/0.96   &\hspace{-9pt} 0.78/1.88   &\hspace{-9pt} 0.59/1.43  &\hspace{-9pt} 0.34/0.86          &\hspace{-9pt} \textbf{0.32}/\textbf{0.84} \\ 
\cmidrule(lr){0-6}
AVG     &\hspace{-9pt} 4.66/7.90     &\hspace{-9pt} 0.89/2.07   &\hspace{-9pt} 0.97/2.27   &\hspace{-9pt} 0.76/1.81  &\hspace{-9pt} 0.62/1.48          &\hspace{-9pt} \textbf{0.48}/\textbf{1.31} \\ 
\bottomrule
\end{tabular}
}
\end{table}

We have further collected two new heterogeneous datasets, UTOWN and CROSS, which contain various types of traffic agents such as pedestrians, cars, bicycles, and buses. 
The datasets can be downloaded from \texttt{\url{https://github.com/AdaCompNUS/GAMMA}}.
Our baselines include SRLSTM, the best-performing \emph{data-driven} baseline for homogeneous datasets, and PORCA, the best-performing \emph{traditional} baseline for homogeneous datasets. We also compare GAMMA with TrafficPredict, which is designed specifically for heterogeneous traffic agents. SGAN and SGAN-ave are also added for comparison.

\tabref{tab:traditional-approaches} shows ADE and FDE for all algorithms. \modelname significantly outperforms other models, especially for the challenging CROSS dataset, where heterogeneous traffic agents interact intensively at a cross in China. See the supplementary material or 
\textit{\url{https://www.dropbox.com/s/v2qqr94520kzg0e/gamma.mp4?dl=0}}
for a video of prediction results on the CROSS dataset.


\subsubsection{Ablation Study}
\label{sec:ablation-study}
\begin{table}[t]
\centering
\caption{Ablation study for \modelname on CROSS. ADE/FDE are reported.\vspace{-5pt}}\label{tab:ablation-study}
{\scriptsize
\begin{tabular}{cccc}
\toprule
\multicolumn{1}{c}{\begin{tabular}[c]{@{}c@{}}w/o kinematic \\constraints\end{tabular}}  \hspace{-5pt}& \multicolumn{1}{c}{\begin{tabular}[c]{@{}c@{}}w/o polygon\\ representation\end{tabular}}  \hspace{-5pt}& \multicolumn{1}{c}{\begin{tabular}[c]{@{}c@{}}w/o intention \\inference\end{tabular}}  \hspace{-5pt}& \modelname         \\ 
\cmidrule(lr){0-3}
0.68/1.84   \hspace{-5pt}& 0.84/2.07  \hspace{-5pt}& 0.96/2.29 \hspace{-5pt}& \textbf{0.64}/\textbf{1.77} \\
\bottomrule
\end{tabular}
}
\end{table}

We further conducted an ablation study on the key components of \modelname by disabling kinematics constrains, polygonal geometry, or intention inference.
Particularly, \modelname w/o kinematics assumes all agents to be holonomic; \modelname w/o polygon representation assumes all agents to be disc-shaped; \modelname w/o intention inference assumes all the agents to continue taking the current velocity. \tabref{tab:ablation-study} presents the results evaluated in the CROSS scene.
These results show that each novel component in \modelname contributes significantly to the performance. 

\subsubsection{Speed Comparison}
\label{sec:speed-comparison}

\begin{table}[!t]
\centering
\caption{Comparison of the execution time (in second) of \modelname  with deep learning models SGAN and SRLSTM. All results are measured on a workstation with Intel(R) Core(TM) i7-7700K CPU and Nvidia GTX 1080 GPU.\vspace{-5pt}}\label{tab:speed-comparison}
{\scriptsize
\begin{tabular}{cccccc}
\toprule
         & \begin{tabular}[c]{@{}c@{}}SGAN \\ ($bs=1$)\end{tabular} & \begin{tabular}[c]{@{}c@{}}SGAN \\ ($bs=64$)\end{tabular} & \begin{tabular}[c]{@{}c@{}}SRLSTM\\ ($bs=1$)\end{tabular} & \begin{tabular}[c]{@{}c@{}}SRLSTM\\ ($bs=4$)\end{tabular} & \modelname   \\ 
\cmidrule(lr){0-5}
run time & 3.15e-3                                                          & 1.80e-3                                                           & 2.32e-3                                                         & 2.19e-3                                                         & \textbf{3.92e-4} \\
speed-up & 1x                                                               & 1.75x                                                             & 1.36x                                                           & 1.44x                                                           & \textbf{8.04x}   \\ 
\bottomrule
\end{tabular}
}
\end{table}
A motion prediction model needs to run sufficiently fast to be useful for simulation and sophisticated planning. We compare the speed of \modelname with two deep learning models, SGAN and SRLSTM. 
\tabref{tab:speed-comparison} shows that \modelname can run much faster than the deep learning models, specifically, 8.04x faster than SGAN.

\subsection{\simname Simulation Performance}
In this section, we perform both qualitative and quantitative study on the performance of \simname in simulating urban crowds on real-world maps.

\subsubsection{Benchmark Scenarios}
\figref{fig:benchmarks}(d-f) shows qualitative simulation results on the benchmark scenarios at real-world locations - Africa, Europe and Asia. Comparison with the real-world scenarios shows that the simulated traffic closely represent the reality. More simulation results can be found in the accompanying video or via \url{https://www.dropbox.com/s/iw3hkkznp4dcj17/tro%20summit.mp4?dl=0}.

\subsubsection{Comparison with Rule-based Simulation}

We additionally compare with rule-based behaviours commonly used in simulators to demonstrate quantitatively the sophistication of \modelname. Particularly, we compare \modelname with time-to-collision (TTC) \cite{luo2019iros}, a reactive model that moves agents along lane center-curves and adapts the agents' speeds according to the predicted time-to-collision. 

\begin{figure}[!t]
\centering                                                              \begin{tabular}{c}                                          \Fbox{\includegraphics[width=0.48\textwidth]{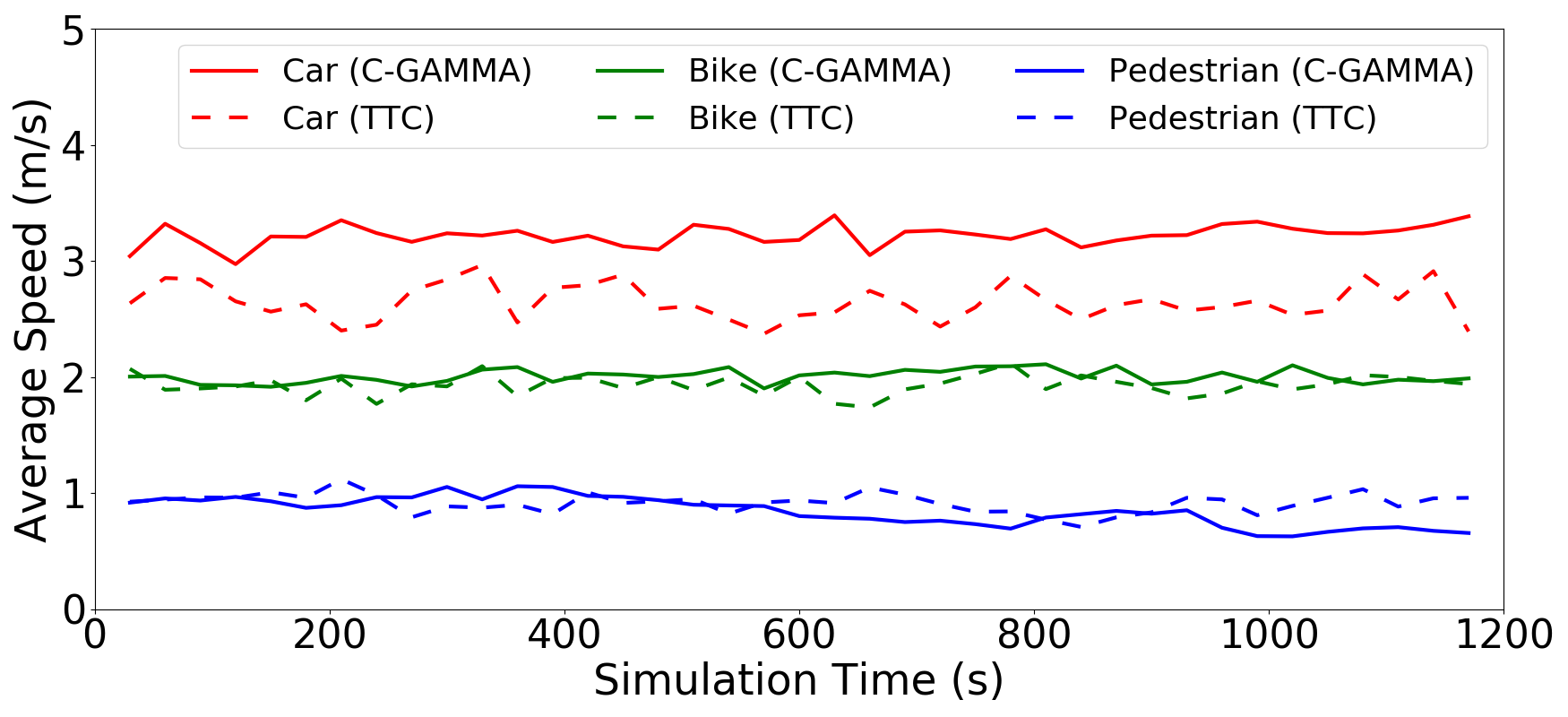}} \\
(\subfig{a}) \\
\Fbox{\includegraphics[width=0.48\textwidth]{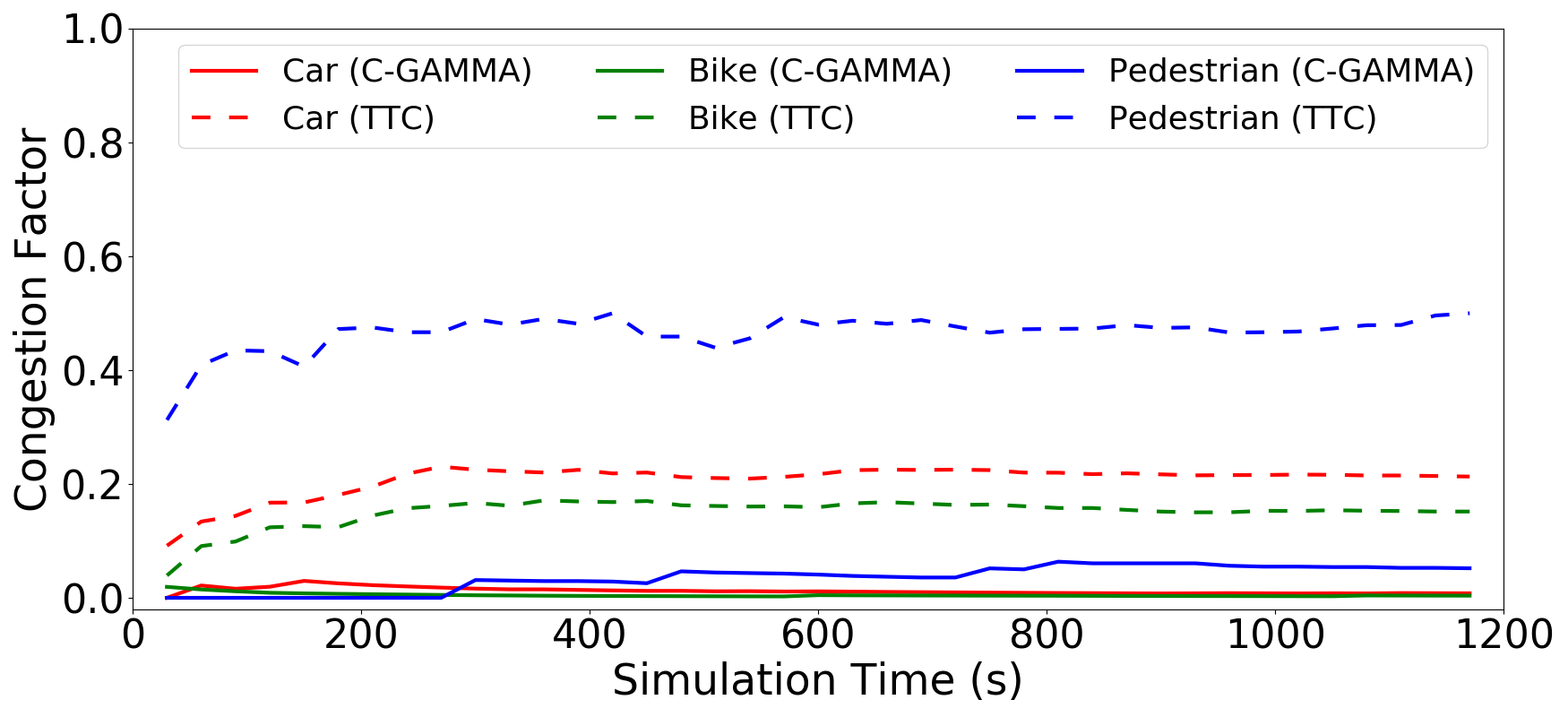}} \\
(\subfig{b})
\end{tabular}
\caption{Performance profile of \modelname and TTC on the Meskel-Intersection: (a) average speed of traffic agents; (b) congestion factor of the traffic.}   \label{fig:simprofile}     
\end{figure}

The criteria for comparison are inspired by human driver's objectives - to maximize driving efficiency and avoid congestion.
We thus measure in our experiments the average speed of traffic agents and a congestion factor defined as the percentage of agents being jammed in the crowd. We remove jammed agents after remaining stationary for a substantial amount of time. 
\figref{fig:simprofile} shows a detailed profile of the agent speeds and the congestion factors for different types of agents against the simulation time. 
\modelname generates faster and smoother traffic than the TTC in all benchmark scenarios throughout 20 minutes of simulation.
The congestion factor of the TTC-controlled traffic grows quickly with the simulation time, indicating that agents fail to coordinate with each other. In contrast, \modelname consistently produces higher agent speeds and low congestion factors for all agent types. 
This is because \modelname explicitly models cooperation between agents and provides an optimal collision avoidance motion using both steering and acceleration. 

\subsection{Performance of the \algname Planner}
We now validate the \algname planner by comparing its driving performance with simpler solutions such as local-collision avoidance and less-sophisticated planning.
For local collision avoidance, we directly use GAMMA to control the ego-vehicle, using the predicted velocities as high-level commands; For simple planning, we use a roll-out algorithm that casts multiple roll-outs following a default policy to choose an optimal immediate  action. The roll-out policy applies the following rules: accelerates the ego-vehicle when exo-agents in front are far way ($>\SI{4}{\meter}$ away), maintains half-speed when they are in caution range ($2\sim\SI{4}{\meter}$ away), and decelerates when they are close-by ($<\SI{2}{\meter}$ away). 

\tabref{tab::plan_performance} shows the performance results including the collision rate per step, the average vehicle speed, and the frequency of deceleration when driving the ego-vehicle using \algname, \modelname, and Roll-out.
In summary, simple planning drives over-conservatively, while local collision avoidance drives over-aggressively. Sophisticated planning using \algname offers a successful trade-off between aggressiveness and conservatives.
Compared to Roll-out that can barely move in the crowd, \algname can drive the vehicle through highly dynamic crowds with significantly higher speed while being safe and smooth.
Compared to \modelname that drives very fast but leads to very high collision rate, \algname achieves similar driving speed with much lower collisions;
We thus conclude that sophisticated long-term planning is important for crowd-driving, and \algname ensures safe, efficient, and smooth drives.

\begin{table}[!t]
\centering
\caption{Comparison on the driving performance of driving algorithms.}
\begin{tabular}{ cccc }
\toprule
&\hspace{-7pt} \thead{Collision / step} &\hspace{-7pt} \thead{Avg. Speed (\si{\Ms})} &\hspace{-7pt} \thead{Dec. / step} \\
\hline
Roll-out & 0.00095 & 2.1 & 0.15 \\
\modelname & 0.002 & 5.53 & 0.12 \\
\algname & 0.00069 & 4.53 & 0.16 \\
\bottomrule
\end{tabular}
\label{tab::plan_performance}
\end{table}

\section{conclusion}
We presented \simname, a simulator for generating high-fidelity interactive data for developing, training, and testing crowd-driving algorithms. 
The simulator uses online maps to automatically construct unregulated dense traffic at any location of the world. 
The core of \simname is a realistic traffic motion model, \modelname, that has been validated on various real-world data sets, and a sophisticated planner, \algname, that achieves safe and efficient driving among dense urban scenes. 
By integrating topological road contexts with the \modelname motion model, \simname can generate complex and realistic crowds that closely represent unregulated traffic in the real-world. 
The \algname planner also serves as a reference planning algorithm for future developments. 
We envision that \simname will support a wide range of applications such as perception, control, planning, and learning for driving in unregulated dense urban traffic.


\bibliographystyle{plain}
\bibliography{main}
\end{document}